\documentclass{article} 

\usepackage[preprint]{colm2026_conference}

\usepackage{microtype}
\usepackage{hyperref}
\usepackage{url}
\usepackage{booktabs}

\usepackage{lineno}

\definecolor{darkblue}{rgb}{0, 0, 0.5}
\hypersetup{colorlinks=true, citecolor=darkblue, linkcolor=darkblue, urlcolor=darkblue}

\usepackage{amssymb}
\usepackage{graphicx}
\usepackage{bm}

\usepackage{fourier} 
\usepackage{array}
\usepackage[export]{adjustbox}
\usepackage{import}
\usepackage{float}
\usepackage{amsmath}
\usepackage{siunitx}
\usepackage{framed}
\usepackage{enumitem}
\usepackage{caption}
\usepackage{footnote}
\usepackage{footmisc}

\DeclareUnicodeCharacter{03A3}{\ensuremath{\Sigma}}

\usepackage{comment}
\usepackage{pifont}
\usepackage{tocloft}

\setlist[2]{noitemsep} 
\setitemize{noitemsep} 
\setenumerate{noitemsep} 

\definecolor{darkblue}{rgb}{0, 0, 0.5}
\hypersetup{colorlinks=true, citecolor=darkblue, linkcolor=darkblue, urlcolor=darkblue}

\title{Key-Value Means: Transformers with Expandable Block-Recurrent Compressed Memory}

\author{Daniel Goldstein\thanks{Equal contribution.} \\
Featherless AI \\
Eleuther AI \\
\texttt{dan@featherless.ai} \\
\And
Navneel Singhal\footnotemark[1] \\
Independent Researcher \\
\texttt{}
\And
Eugene Cheah\\
Featherless AI \\
Eleuther AI\\
\texttt{eugene@featherless.ai}
}

\begin{document}

\ifcolmsubmission
\linenumbers
\fi

\maketitle

\begin{abstract}
Recall presents a difficult choice: transformers have a linearly growing memory that slows each successive token, while linear RNNs typically have fixed costs but limited recall.
We present Key-Value Means ("KVM"), a novel block-recurrence for attention that can accommodate either fixed-size or growing state. Equipping a strong transformer baseline with fixed-size KVM attention layers yields a strong $O(N)$ chunked RNN, while adding only an insignificant number of new parameters. We train a transformer with a growable KVM cache and show it performs competitively on long-context tests with only subquadratic prefill time and sublinear state growth.
KVM is implementable with standard operations and without custom kernels, and supports chunk-wise parallelizable training and prefill. It provides many of the benefits of both traditional transformers (expandable context memory, chunk-wise parallelizable training and prefill) and RNNs in a single unified package. It can be used on every layer, saving KV-cache memory, and allowing a continuous range of choices of prefill time complexity between $O(N)$ and $O(N^2)$.
We release our code \href{https://github.com/featherless-ai/KVM-paper}{here} and trained models \href{https://huggingface.co/collections/featherless-ai/kvm-paper}{here} under the Apache 2.0 license.
\end{abstract}

\section{Introduction} \label{sec:introduction}

Transformers~\citep{vaswani2017attention} are efficient on modern hardware and retain one key-value pair for every previous token, but their memory and time per output token grow linearly with context length. Most modern linear RNNs (LRNNs) use constant memory and time per token, but their fixed-size state typically limits long-context memory. A model that can vary its state growth between these endpoints would provide a selectable trade-off between memory, prefill time, decode time, and recall.

Growable state is useful for recall-heavy tasks in which potentially relevant difficult-to-compress information accumulates over the context. An architecture with fixed state size must represent an increasing amount of such information with unchanged capacity, whereas a growable state can allocate additional capacity as the context length increases, despite remaining smaller than a full KV cache. An architecture whose state grows more slowly than its context must represent multiple tokens using fewer state entries, and the difficulty of doing so depends on the redundancy of the context.

\begin{figure}[H]
    \centering
    \includegraphics[width=0.6\linewidth]{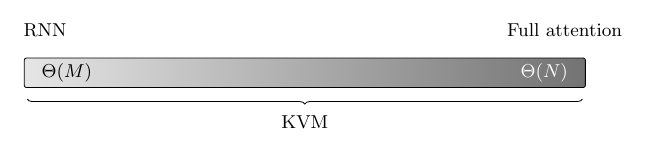}
    \caption{KVM state growth interpolates between fixed-state RNNs and full transformer KV cache.}
    \label{fig:interpolating}
\end{figure}

Key-Value Means (KVM) uses softmax attention across an uncompressed block sliding window and a compressed state, whose size can remain fixed or grow according to a flexibly chosen budget. This gives us a chunked recurrent architecture whose memory and computation can be selected across a continuous range. Figure~\ref{fig:interpolating} and Table~\ref{tab:comparison} summarize these trade-offs.

\begin{table}[ht]
\centering
\begin{adjustbox}{max width=0.75\linewidth}
\begin{tabular}{lcccc}
Property & Fixed-state RNN & KVM (fixed) & KVM ($\sqrt{N}$) & Full Attn \\
\midrule
State Size & $O(1)$ & $O(1)$ & $O(\sqrt{N})$ & $O(N)$ \\
\midrule
Prefill time per sequence & $O(N)$ & $O(N)$ & $O(N^{1.5})$ & $O(N^2)$ \\
\midrule
Decode time per token & $O(1)$ & $O(1)$ & $O(\sqrt{N})$ & $O(N)$ \\
\midrule
Parallel prefill & Usually chunk-wise & Chunk-wise & Chunk-wise & Fully \\
\bottomrule
\multicolumn{5}{c}{$N = \text{sequence length}$}
\end{tabular}
\end{adjustbox}
\caption{KVM: Interpolating between RNNs and Transformers}
\label{tab:comparison}
\end{table}

Our main contributions are:

\begin{itemize}
\item Key-Value Means, a high performance block-recurrent softmax-attention layer that combines an exact block sliding window with a growable compressed KV state to provide a fast, low state-size alternative to full attention, and higher recall alternative to fixed state-size RNN layers.
\item A method of applying a pre-selected growth strategy to interpolate between RNNs and Transformers.
\item A method of sharing partial RoPE and NoPE across compressed and uncompressed state regions in single softmax attention without significantly increasing parameters.
\item A just-in-time (JIT) key renormalization scheme for compressed states.
\item A method of value radius preservation.
\item Optimized open-source kernels for fast training, prefill, and decoding of Key-Value Means layers.
\end{itemize}

\section{Background} \label{sec:background}

Architectural methods that grow less rapidly than transformers linearly growing KV cache must make several choices about how to achieve this reduced growth:

\paragraph{Compression}
Many state compression techniques have been proposed for auto-regressive transformers and parallelizable alternatives, with popular ones falling into several classes: LRNNs \citep{katharopoulos2020lineartransformrers, peng2025rwkv7gooseexpressivedynamic, yang2025gated} typically employ a fixed size state updated at every new token. Sparsely routed LRNNs \citep{behrouz2026memory, afzalbick2026raven, cabannes2026sparsedeltamemoryscaling} own several such states and choose one or more of them at each new token for update and retrieval. State test-time training ("TTT") \citep{sun2025learning, zhang2026testtime, behrouz2025titans} methods are the run-time analogue to pre-training, with a state update that fulfills some optimization criteria, e.g. least-squares/SGD on an MLP-shaped state to reduce the difference between each value and its key with the state applied to it. Many modern LRNNs (e.g. RWKV-7, GDN) fall into this category. Non-LRNN TTT architectures commonly use batched rather than per-token updates. KV eviction \citep{NEURIPS2023_6ceefa7b, 2023keyformer} strategies start with a standard transformer and choose which tokens are removed from the KV cache. State Merging \citep{liu2025zsmergezeroshotkvcache, alonso2026onlinevectorquantizedattention} strategies typically keep a KV cache shaped state, and decide which sets of tokens get merged together to form a smaller set. Many techniques combine eviction with merging, with evicted tokens being conditionally merged into a compressed segment of the state. 

\paragraph{Timing}
In order for new tokens to be admitted into consideration for output, the recurrent state of an auto-regressive model must be updated for each new token. In a standard transformer, each token is simply appended to the KV cache. Sliding Window Attention \citep{xiao2024efficientstreaminglanguagemodels} is one of the simplest eviction methods, treating state as a fixed length queue. Block sliding window attention \citep{hwang2024transformerfamfeedbackattentionworking} also acts as a queue, but periodically evicts an entire tail block at a time when full.

Although it must be updated in some manner, the state need not be updated in the same way for each incoming token. Many post-training KV cache compression methods append every token during prefill, but transition to do some combination of eviction and compression afterwards. This can be done as a one-time, once-per-token, or periodic update during decoding.

LRNNs can update their fixed-size state once for every incoming token. However, TTT methods that feature larger/deeper or otherwise more complex states often use a batched update in order to amortize update cost. Such batching is an elegant solution, but the state must still be somehow updated for every incoming token. Typical fixes are to add a separate SWA region and combining gate, or place the TTT on a separate layer alternating with SWA layers. Another less common fix is to employ BSWA and have the tail block update the TTT state upon eviction.

\paragraph{Compression Metrics}
Methods that treat eviction or merging as a choice need some metric for the decision. Many post-hoc KV cache compression methods rely on some form of measurement of instantaneous or cumulative attention mass to make this decision. Others compare key or even value similarity.

\paragraph{Positional Encoding}

Methods that merge state must make decisions about how to apply any positional encoding during the merge operation. LRNNs often use some form of decay or forget gate to encode position rather than explicit encodings.

\paragraph{Our Choices}
From this palette of options, KVM chooses a unique set: BSWA with evicted tokens appended/merged into a compressed state in batched fashion, selectable state growth rate, token radius tracking and Just-In-Time renormalization, a metric based on maximum key-similarity, and rotary positional information removed from keys as they are merged into the state. As we will demonstrate below, these choices result in a unified layer architecture, competitive speed profile, and strong modeling and recall performance even at reduced state size.

Please see Appendix \ref{sec:related-work} for a detailed discussion of related work.

\section{Method}\label{sec:method}

\subsection{Overview}

\paragraph{Compressed State and BSWA Window}

KVM processes a sequence in blocks. It keeps the keys and values of the most recent blocks in an uncompressed Block Sliding Window Attention (BSWA) window, and stores information from earlier blocks in a compressed key-value state. The state contains a set of key-value rows, and the state size may remain fixed or grow according to a chosen state budget. A fixed set of initial state rows is protected from later updates as attention sinks. At each block boundary, the oldest block leaves the BSWA window and updates the state. Queries attend with a single softmax over the concatenation of the state and the BSWA window. This also allows for chunk-wise parallel training and prefill.

\paragraph{State Compression and JIT Normalization}

For each token to be merged from a BSWA overflow block, KVM compares its normalized key with the normalized keys of the eligible (non-sink) state rows and assigns it to the row with the largest similarity. The token key and value are multiplied by a learned merge gate and added to the running key and value sums in the selected state row. Immediately before the state is used for attention, KVM applies LayerNorm to each state key and restores each state value to the radius stored when the row was created. This just-in-time normalization prevents norm shrinkage as vectors are combined and removes the need to track a count for each state row.

\paragraph{State Initialization and Expansion}

A fixed or growing state budget determines how many overflow tokens are appended to the state. At the first state-creation step, the state is initialized by directly copying the overflow tokens. When the state budget grows, KVM appends the overflow tokens with the lowest maximum similarity to the existing state rows. The remaining overflow tokens are then merged using the compression rule above. If the state budget does not grow, all overflow tokens are merged, and similarly, if the state budget allows it, all overflow tokens are appended.

\paragraph{Positional Encoding}

The compressed state stores global information from blocks preceding the short uncompressed BSWA window, and avoids dealing with positional information of the input keys. KVM applies partial RoPE to queries and keys in the BSWA window, and zeros the rotation-relevant subspace of keys before merging them into the compressed state. Concatenating these permits a single attention call over the state and BSWA window without additional projection matrices.

Appendix~\ref{sec:design-rationale} discusses the rationale and alternatives for these design choices.

\subsection{KVM Formulation}

\paragraph{Terminology and notation}

We use \emph{state} for key-value information (keys, values, and radii) carried from one chunk to later chunks. Each key-value pair is a \emph{state row}, or memory slot, and the state size is its number of rows. We use \emph{block} and \emph{chunk} interchangeably for a contiguous group of tokens. The Block Sliding Window Attention (BSWA) window retains several recent uncompressed blocks, and an \emph{overflow block} is the oldest block as it leaves this window.

For a sequence of length $T$, $\mathbf{x}_t\in\mathbb{R}^d$ denotes the residual-stream vector at sequence position $t$. We write $\mathbf{q}_t$, $\mathbf{k}_t$, and $\mathbf{v}_t$ for the query, key, and value of one attention head, whose dimension is $d_h$, and $H$ denotes the number of heads. State rows use the index $i$.

\begin{figure}
    \centering
    \includegraphics[width=1\linewidth]{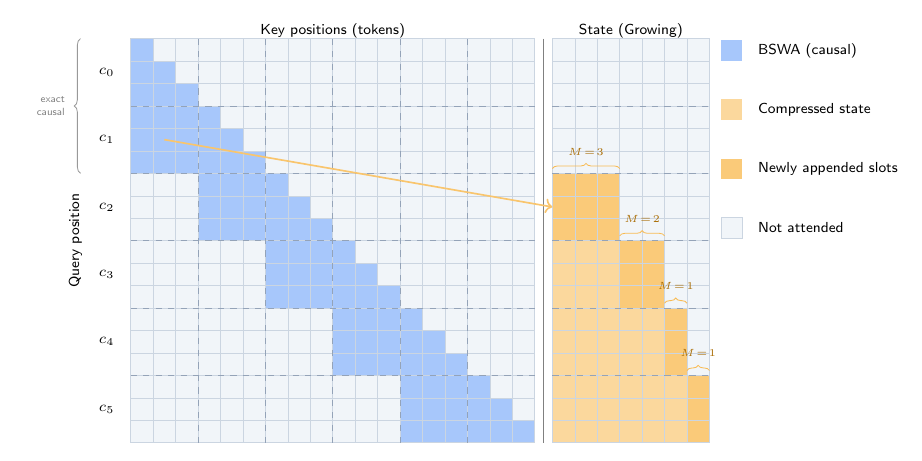}
    \caption{\centering{KVM Attention mask across both causal BSWA and growing KVM state} \\ 
    \footnotesize{$C=3$, $\texttt{n\_bswa\_chunks}=2$, window $L\!=\!6$}}
    \label{fig:attn-mask}
\end{figure}

KVM applies traditional softmax attention over keys and values from (1) a fixed set of StreamingLLM-style sink tokens~\citep{xiao2024efficientstreaminglanguagemodels}, (2) the uncompressed BSWA window~\citep{hwang2024transformerfamfeedbackattentionworking}, and (3) the periodically updated and JIT-normalized state. In the implementation and equations below, the sink tokens are protected rows of the state, although they could be implemented separately. At the end of each block, KVM appends zero or more tokens from the overflow block to the state and merges the remaining overflow tokens. The append and merge operations are defined below. Figure~\ref{fig:attn-mask} shows the attention regions, and Appendix~\ref{sec:pseudocode} gives pseudocode.

\paragraph{Chunks and attention regions}

Let $C=\texttt{chunk\_len}$ and $L=\texttt{n\_bswa\_chunks}\cdot C$. The first $L_0=\min(T,L)$ tokens use exact causal attention over the available prefix, with regional temperatures $\tau_{\mathrm{state}}$, $\tau_{\mathrm{bswa}}$ described below. After that, KVM processes one chunk $[s,e)$ of query tokens at a time. For a chunk $[s,e)$, define the beginning of the BSWA window as $b=e-L$. Subscripts $t$ and $i$ denote sequence position and state position, respectively. We consider a single head for notational convenience. See Appendix \ref{sec:gptalpha} for details on the overall transformer architecture used in our experiments.

\begin{figure}
    \centering
    \includegraphics[width=0.875\linewidth]{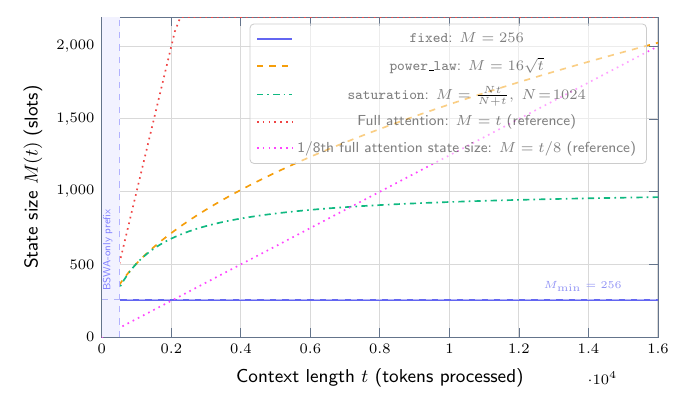}
    \caption{Examples of fixed, power-law, and saturating KVM state-budget schedules.}
\end{figure}

\paragraph{KVM weight preparation}
To make the state independent of the positions of the original tokens, KVM zeros the rotary subspace of the overflow keys (the first $r$ channels out of a total of $d_h$ head channels) and normalizes keys using a standard LayerNorm with bias before their use as memory keys. The merge gate, a scalar for each head calculated from the incoming $x_t$, modulates the amount of each incoming overflow key that the state will absorb, in a data-dependent fashion.

\begin{align}
\bar{\mathbf{k}}_t &= \operatorname{LN}_s(\mathbf{k}_t \cdot \operatorname{diag}(\underbrace{0,\ldots,0}_{r},\underbrace{1,\ldots,1}_{d_h-r})) &\text{memory key}\\
g_t &= 1+\operatorname{ELU}(\mathbf{x}_t W_g), \quad W_g\in\mathbb{R}^{d \times 1} &\text{merge gate}\\
\breve{\mathbf{k}}_t &= {g_t}\,\bar{\mathbf{k}}_t &\text{gated memory key}\\
\breve{\mathbf{v}}_t &= {g_t}\,\mathbf{v}_t &\text{gated value}
\end{align}

The initial state is always one chunk long, and is formed from the first chunk of \(\bar{\mathbf{k}}\) and \(\mathbf{v}\). The first chunk initializes the state and is not later processed as an overflow block. $\rho_i$ stores the value readout radius of state row $i$, and remains static throughout its lifetime. $m$ is the current number of state rows, initially equal to $C$. For each $i\in[0,m)$,

\[
\mathbf{s}_i^K = \bar{\mathbf{k}}_i, \quad \mathbf{s}_i^V = \mathbf{v}_i, \quad \rho_i=\|\mathbf{s}_i^V\|_2
\]

The query $\mathbf{q}$ has been token shifted, normalized and partially RoPE-rotated by this point, per the GPTAlpha-2 weight preparation in Appendix \ref{sec:gptalpha}. 

\paragraph{Readout}
Before attention, the state is temporarily normalized row-wise:
\[
\hat{\mathbf{s}}_i^K=\operatorname{LN}_s(\mathbf{s}_i^K),
\qquad
\hat{\mathbf{s}}_i^V=\rho_i\frac{\mathbf{s}_i^V}{\max(\|\mathbf{s}_i^V\|_2, \epsilon_{\text{norm}})}
\]
where \(\epsilon_{\mathrm{norm}}>0\) is a small numerical stabilizer.
KVM then attends to the concatenation of the normalized state and the unchanged BSWA window:
\[
K^A=
\begin{bmatrix}
\tau_{\mathrm{state}}\,\hat{\mathbf{s}}_{0:m}^K\\
\tau_{\mathrm{bswa}}\,\mathbf{k}_{b:e}
\end{bmatrix},
\qquad
V^A=
\begin{bmatrix}
\hat{\mathbf{s}}_{0:m}^V\\
\mathbf{v}_{b:e}
\end{bmatrix}
\]
where $\tau_{\mathrm{state}}$, $\tau_{\mathrm{bswa}}$ are learned per-head scalar inverse temperatures. For each query row $u\in[s,e)$,
\[
\mathbf{y}_u
=
\operatorname{softmax}\!\left(
\frac{{\mathbf{q}}_u(K^A)^\top}{\sqrt{d_h}}+\mathbf{M}_u
\right)V^A
\]
where $\mathbf{M}_u$ leaves all state rows visible and applies causal masking within the BSWA window.

Then, as usual, per-head outputs are concatenated and projected back to $\mathbb{R}^d$:
\[
\mathbf{y}_t=\operatorname{Concat} \bigl(\mathbf{y}_t^{(1)},\ldots,\mathbf{y}_t^{(H)}\bigr)W_O,
\qquad
W_O\in\mathbb{R}^{d\times d}
\]
and the result is added to the residual stream.

\paragraph{KVM Recurrence}

\paragraph{Append}

At the end of each chunk, one chunk of overflow tokens falls off the back of the BSWA window. Let \(\Omega_e=[b,b+C)\) denote the overflow block incorporated into the state after attending to queries for chunk \([s,e)\). If $n_{\mathrm{append}}>0$ (which we specify later), we append the $n_{\mathrm{append}}$ least redundant overflow tokens to the state, where redundancy is measured against the current normalized state. For each $j \in \Omega_e$,
\[
s_j=\max_i\, \bar{\mathbf{k}}_j\hat{\mathbf{s}}_i^{K\top}\]

Let $A_e\subseteq\Omega_e$ be the $n_{\mathrm{append}}$ indices with the smallest scores $s_j$. These tokens are appended directly:
\[
\mathbf{s}_+^K=
\begin{bmatrix}
\mathbf{s}_{0:m}^K\\
\bar{\mathbf{k}}_{A_e}
\end{bmatrix},
\qquad
\mathbf{s}_+^V=
\begin{bmatrix}
\mathbf{s}_{0:m}^V\\
{\mathbf{v}}_{A_e}
\end{bmatrix},
\qquad
\boldsymbol{\rho}_+=
\begin{bmatrix}
\boldsymbol{\rho}_{0:m}\\
\|{\mathbf{v}}_{A_e}\|_2
\end{bmatrix}
\]

where \(\|\mathbf{v}_{A_e}\|_2\) is taken row-wise.

\paragraph{Merge}

The remaining overflow tokens $R_e=\Omega_e\setminus A_e$ are then merged into the \emph{updated} state $\mathbf{s}_+$. (The merge targets include both rows that existed previously as well as any rows appended in the same step.) 
The first $S = 1$ state rows are protected as sinks and cannot be selected as merge targets. For each token $j$ to be merged, the merge target $\pi_e(j)$ is given by:

\[
\pi_e(j)
=
\operatorname*{arg\,max}_{i\ge S}\,
\breve{\mathbf{k}}_j \operatorname{LN}_s(\mathbf{s}_{+,i}^{K})^\top
=
\operatorname*{arg\,max}_{i\ge S}\,
\bar{\mathbf{k}}_j \operatorname{LN}_s(\mathbf{s}_{+,i}^{K})^\top
\]
The merge update is, for each state token $i$,
\[
\mathbf{s}_{\text{new},i}^{K} = \mathbf{s}_{+,i}^K + \sum_{j : \pi_e(j) = i} \breve{\mathbf{k}}_j, \quad \mathbf{s}_{\text{new,}i}^{V} = \mathbf{s}_{+,i}^V + \sum_{j : \pi_e(j) = i} \breve{\mathbf{v}}_j
\]

We choose $n_{\text{append}}$ as follows. Suppose $\mathcal{B}(e)$ is the state budget in terms of number of state tokens that we wish to use for the next chunk - e.g., it can be a constant, power-law, or saturating function. Our desired state size is non-decreasing, and we denote it by $M^\star(e)=\max\,\Bigl(m,\;\min\bigl(\mathcal{B}(e),\, m + C\bigr)\Bigr)$. The number of tokens we wish to append is $n_{\mathrm{append}}=\min\!\bigl(M^\star(e)-m,\,|\Omega_e|\bigr)$. Here, $m + C$ caps the budget to not overflow beyond the available number of tokens (state plus overflow tokens), and $m$ bounds the budget from below to avoid removing state rows.

Note that the radii $\rho_i$ are updated only when a slot is \emph{created}, and remain static for the slot thereafter. At readout, the value state is always renormalized back to the stored radius,
\[
\hat{\mathbf{s}}_i^V=\rho_i\frac{\mathbf{s}_i^V}{\max(\|\mathbf{s}_i^V\|_2, \epsilon_{\text{norm}})}.
\]
So merging tokens into the state changes the \emph{direction} of $\hat{\mathbf{s}}_i^V$, while the norm used at this readout remains fixed at the slot's stored radius. This was motivated by the observation that sink tokens in standard attention have small value vector magnitudes (\citet{guo2024attentionscoreneedtoken}).  We experimented with combining norms of value vectors of tokens assigned to the current slot, but did not observe any added benefit on top of this. 

Note that we perform normalization before readout for $\mathbf{s}^K$ / $\mathbf{s}^K_+$ as well. The effect of doing so is equivalent to taking the weighted mean (weights defined using ${g}_j$) of tokens assigned to the slot and then mapping to a shifted hyperellipsoid.

\section{Language modeling performance}\label{lm-evaluations}

To demonstrate the relative performance of the KVM architecture in various configurations, we train a series of models at 120M and 350M parameters for 3B and 7.8B tokens respectively on the Prolong dataset \citep{gao2025prolong} at 8k context length. We choose RWKV-7 \citep{peng2025rwkv7gooseexpressivedynamic} as a strong LRNN baseline \citep{hu2025improving, cao2026optimaldecayspectralinear}. KVM variants use block size $C = 256$ and $\mathtt{n\_bswa\_chunks} = 2$, the former giving us approximately the best prefill and update speed in both PyTorch and kernel experiments with good resulting model quality. "KVM 256" has a fixed compressed state of 256 tokens; "KVM sqrt" uses a $16\sqrt{N}$ state growth schedule. All models share the GPTAlpha-2 backbone described in Section \ref{sec:gptalpha}, with the exception of RWKV-7 for which we use the RWKV-7 backbone; hybrid variants interleave a 1024 saturating state scheduled KVM or Online Vector Quantized Attention(OVQ) \citep{alonso2026onlinevectorquantizedattention} with 256-token RoPE-based SWA on alternate layers. "GPTA" is a pure GPTAlpha-2 model with full attention on every layer and RoPE applied on half of the head channels (called HalfRoPE). "BSWA" is a pure Block Sliding Window Attention model with three blocks and the same half RoPE. For the hybrid GPTA/SWA model, we train two full-attention layer varieties: a half RoPE version, and a NoPE version. Please see Appendix \ref{sec:hyperparameters} for training details.

\paragraph{Loss over sequence position} We evaluate our 120M/350M models by computing mean loss over blocks of size 1024 tokens on a random subset of TextbookChapters \citep{chevalier2024language} documents of length at least 32768 tokens. We observe that KVM has strong performance as the sequence position increases. Notably, even the fixed state size KVM 256 outperforms the much larger state OVQ/SWA (saturating schedule) in this test. Note that KVM-sqrt displays the best results of any non-GPTAlpha model tested.

During experimentation we observed interesting interactions between the variety of RoPE, training token count, and extrapolation performance. Please see Appendix \ref{sec:extrapolation} for details and experiments. KVM and OVQ both eschew positional information entirely on their state, but because of the way KVM works it is able to use relative positional information on its BSWA region. When considered in the context of our RoPE ablation results, it seems that this may be one cause for KVM's larger performance gains in extrapolation versus OVQ.

\begin{figure}
    \centering
    \includegraphics[width=1\linewidth]{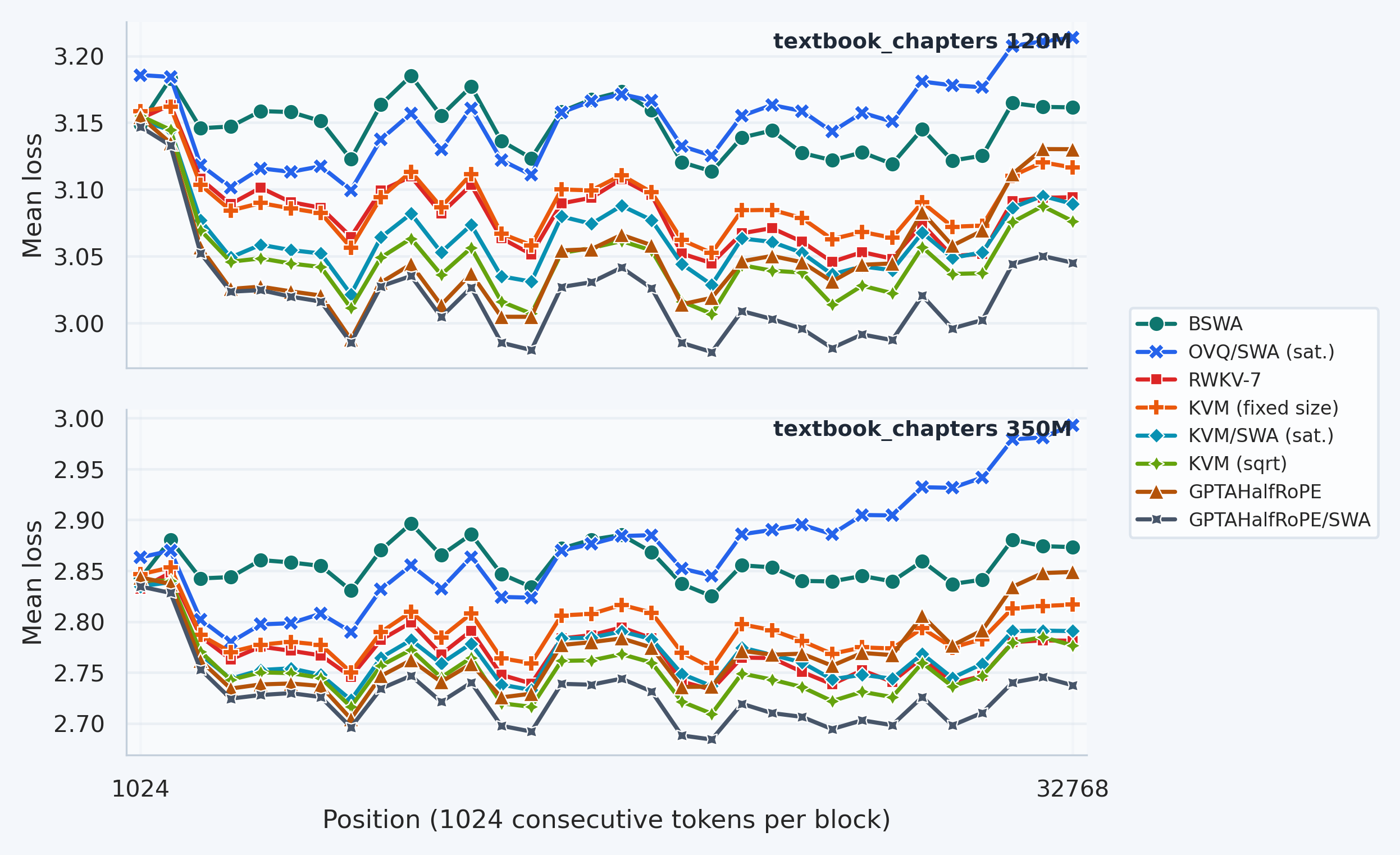}
    \caption{TextbookChapters mean loss per 1024 token block}
    \label{fig:textbook-chapters}
\end{figure}

\paragraph{Standard short-context benchmarks}

Because KVM naturally attends jointly over the BSWA window and the compressed state due to its design, it should behave similarly to a standard transformer on tasks contained within the BSWA window. Our window is such that it fits many standard short-context benchmark tasks. We test KVM and other architectures on various standard short-context benchmarks using LM Evaluation Harness \citep{eval-harness}, and find that results are consistent with this expectation. For experimental results and comparison please see Appendix \ref{sec:evals-short}.

\paragraph{RULER \citep{hsieh2024ruler} and LongBench \citep{bai2024longbenchbilingualmultitaskbenchmark}}

To evaluate the long-context capabilities of KVM and other architectures, we evaluate the 120M/350M models on the NIAH-S subset of RULER at various context lengths (to evaluate both trained context length and out-of-the-box extrapolation), full RULER at 4k context length, and the few-shot subset of LongBench, all using LM Evaluation Harness \citep{eval-harness}. We report our findings in Table~\ref{tab:evals-long}. Unlike in the loss over sequence position experiments above, here we see that KVM-256 has difficulties at extremely long context length in NIAH-S2 and NIAH-S3, but that KVM-sqrt and KVM-sat/SWA hybrid perform well. These specific NIAH variants use a long essay as distractor instead of repeated text. This poses a challenge for any model with a small state size, including RWKV-7. Such models are able to effectively ignore repeated distractors by reusing state entries, but such a strategy becomes untenable when those distractors are continuously novel. This suggests that the ability to utilize increasing state size can be a significant benefit.

\begin{table}[htb]
    \centering
    \begin{adjustbox}{max width=\linewidth}
        \begin{tabular}{l*{14}{r}}
            \toprule
            & \multicolumn{4}{c}{NIAH-S1$\uparrow$}
            & \multicolumn{4}{c}{NIAH-S2$\uparrow$}
            & \multicolumn{4}{c}{NIAH-S3$\uparrow$} & LB$\uparrow$ & RULER$\uparrow$ \\
            \cmidrule(lr){2-5} \cmidrule(lr){6-9} \cmidrule(lr){10-13}
            Architecture & 4K & 8K & 16K & 32K & 4K & 8K & 16K & 32K & 4K & 8K & 16K & 32K & avg. & avg. \\
            \midrule
            120M BSWA & 18.4 & 8.2 & 3.8 & 2.6 & 19.0 & 10.2 & 4.8 & 2.4 & 17.4 & 6.4 & 6.6 & 2.6 & 11.7 & 9.4 \\
            120M RWKV-7 & 97.2 & 95.4 & 71.6 & 9.8 & 4.4 & 1.6 & 0.4 & 1.0 & 4.8 & 2.0 & 1.4 & 0.2 & 17.5 & 15.6 \\
            120M GPTA-2 & 100.0 & 99.6 & 87.8 & 29.6 & 99.8 & 99.2 & 32.8 & 8.4 & 59.4 & 26.0 & 2.4 & 3.8 & 16.9 & 34.0 \\
            120M KVM 256 & 99.4 & 97.8 & 98.4 & 98.4 & 88.8 & 44.0 & 2.6 & 2.4 & 27.2 & 2.0 & 1.2 & 2.6 & 12.2 & 25.2 \\
            120M KVM sqrt & 100.0 & 99.8 & 99.8 & 99.6 & 93.8 & 52.4 & 19.0 & 4.2 & 65.0 & 43.4 & 16.4 & 2.6 & 16.6 & 29.6 \\
            \midrule
            120M OVQ/SWA & 99.8 & 81.8 & 46.2 & 22.2 & 27.6 & 24.4 & 3.6 & 2.0 & 20.8 & 18.6 & 5.0 & 2.4 & 12.0 & 16.5 \\
            120M GPTA-2 HalfRoPE/SWA & 100.0 & 100.0 & 75.0 & 31.4 & 99.8 & 99.6 & 43.2 & 12.4 & 91.8 & 81.6 & 26.2 & 2.0 & 12.2 & 38.7 \\
            120M GPTA-2 NoPE/SWA & 100.0 & 100.0 & 100.0 & 99.6 & 99.4 & 94.8 & 51.0 & 2.4 & 46.2 & 72.2 & 4.0 & 0.0 & 10.0 & 30.9 \\
            120M KVM/SWA & 97.0 & 94.8 & 50.4 & 38.6 & 98.2 & 89.6 & 9.4 & 2.8 & 20.0 & 8.8 & 0.8 & 2.6 & 12.6 & 27.5 \\
            \midrule
            350M BSWA & 18.6 & 8.2 & 3.8 & 2.6 & 19.0 & 10.2 & 4.8 & 2.4 & 19.0 & 5.2 & 5.6 & 2.4 & 17.3 & 12.0 \\
            350M RWKV-7 & 99.6 & 98.2 & 93.8 & 12.6 & 21.6 & 7.0 & 2.0 & 3.4 & 3.0 & 0.6 & 1.2 & 0.4 & 23.2 & 19.6 \\
            350M GPTA-2 & 100.0 & 100.0 & 51.8 & 22.4 & 100.0 & 99.8 & 45.2 & 20.8 & 98.2 & 72.2 & 37.6 & 5.8 & 25.1 & 47.2 \\
            350M KVM 256 & 99.0 & 99.8 & 99.2 & 98.8 & 99.0 & 75.6 & 3.6 & 2.4 & 69.6 & 18.6 & 3.0 & 2.4 & 23.7 & 33.2 \\
            350M KVM sqrt & 100.0 & 99.2 & 99.2 & 98.4 & 98.8 & 97.2 & 71.0 & 17.0 & 97.6 & 95.0 & 77.6 & 36.6 & 25.0 & 38.6 \\
            \midrule
            350M OVQ/SWA & 99.4 & 98.2 & 89.8 & 45.2 & 95.8 & 51.2 & 17.6 & 4.2 & 59.6 & 32.4 & 8.6 & 2.6 & 21.0 & 29.0 \\
            350M GPTA-2 HalfRoPE/SWA & 99.2 & 99.6 & 92.6 & 41.2 & 99.4 & 99.4 & 52.0 & 23.6 & 88.2 & 60.4 & 37.2 & 9.6 & 13.6 & 41.6 \\
            350M GPTA-2 NoPE/SWA & 99.4 & 99.6 & 99.6 & 95.0 & 99.4 & 97.6 & 99.0 & 0.0 & 82.2 & 47.8 & 29.2 & 0.0 & 21.0 & 45.1 \\
            350M KVM/SWA & 99.6 & 99.8 & 96.0 & 92.2 & 99.6 & 97.0 & 47.6 & 5.8 & 95.8 & 77.8 & 28.8 & 6.8 & 25.2 & 40.7 \\
            \bottomrule
        \end{tabular}
    \end{adjustbox}
    \caption{NIAH, RULER-4096 and average of LongBench ("LB") few-shot evaluations}
    \label{tab:evals-long}
\end{table}

\section{Speed Benchmarks}

We implement KVM kernels in Triton~\citep{tillet2019triton} and compare them with full causal multi-head attention for autoregressive decode, prefill, and the backward pass.

We benchmark the sequence-mixing operation of a single layer with $32$ heads each of dimension $128$, at batch size $8$, using bfloat16 for inputs and outputs and
float32 for precision-sensitive operations.
We evaluate both KVM-256 and KVM-sqrt on context lengths from 512 to 32768 tokens, on an AMD Instinct MI325X GPU. The attention baseline uses PyTorch's Flash SDPA using ROCm AOTriton kernels.

Timing begins after the input projections and ends after head concatenation. It therefore includes KVM's routing and recurrent-state computation and full attention's causal attention computation, while excluding projections, positional encoding, and the remainder of the transformer layer for both methods.

\begin{figure*}[t]
    \centering
    \includegraphics[width=\textwidth]{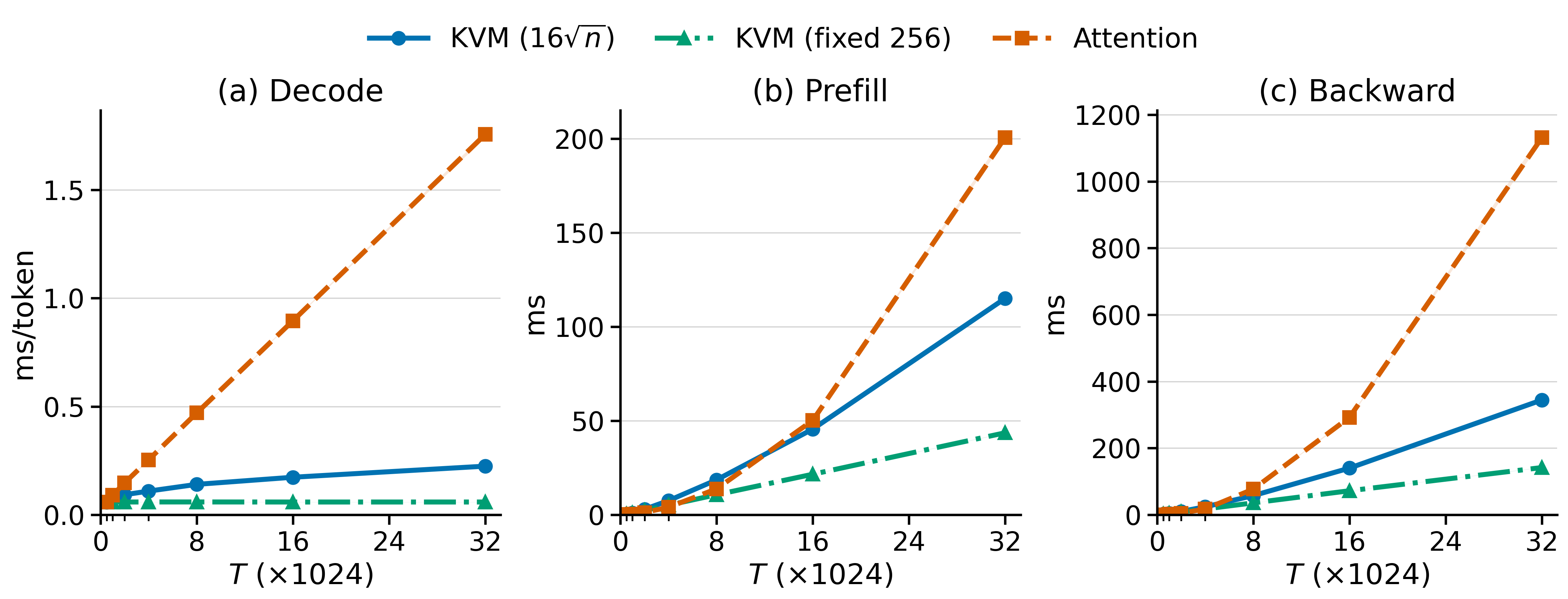}
    \caption{Sequence-mixing time for autoregressive decode (left), prefill
    (center), and backward (right).  Lines show medians and bands span the
    minimum and maximum run medians.  Lower is better.}
    \label{fig:kvm-kernel-benchmarks}
\end{figure*}

Figure~\ref{fig:kvm-kernel-benchmarks} shows an increasing advantage for KVM at longer contexts. At $T=32768$, KVM-256 is $29.77\times$ faster for decode, $4.59\times$ faster for prefill, and $7.99\times$ faster for backward than full attention. KVM-sqrt achieves speedups of $7.88\times$, $1.76\times$, and $3.30\times$, respectively. KVM-256 becomes faster than full attention at 1K tokens for decode, 8K tokens for prefill, and 4K for backward, while KVM-sqrt becomes faster at 1K, 16K, and 8K tokens, respectively.

\section{Ablation studies} \label{ablations}

We run a series of ablation studies to examine the contributions of each part of the KVM architecture, on 120M KVM 256 models. We report long context evals in Table~\ref{tab:ablations-evals-long}, and short-context evals in Table~\ref{tab:ablations-evals-short} in Appendix~\ref{sec:evals-short}.

\begin{table}[htb]
    \centering
    \begin{adjustbox}{max width=\linewidth}
        \begin{tabular}{l*{14}{r}}
            \toprule
            & \multicolumn{4}{c}{NIAH-S1$\uparrow$}
            & \multicolumn{4}{c}{NIAH-S2$\uparrow$}
            & \multicolumn{4}{c}{NIAH-S3$\uparrow$} & LB$\uparrow$ & RULER$\uparrow$ \\
            \cmidrule(lr){2-5} \cmidrule(lr){6-9} \cmidrule(lr){10-13}
            Architecture & 4K & 8K & 16K & 32K & 4K & 8K & 16K & 32K & 4K & 8K & 16K & 32K & avg. & avg. \\
            \midrule
            baseline & 99.4 & 97.8 & 98.4 & 98.4 & 88.8 & 44.0 & 2.6 & 2.4 & 27.2 & 2.0 & 1.2 & 2.6 & 12.2 & 25.2 \\
            no sink & 83.8 & 86.4 & 71.4 & 79.6 & 37.6 & 4.6 & 1.6 & 1.6 & 10.8 & 1.6 & 0.0 & 0.0 & 14.0 & 19.1 \\
            no head temps & 97.4 & 98.4 & 97.8 & 98.8 & 55.0 & 14.6 & 2.2 & 2.4 & 50.8 & 11.2 & 0.6 & 2.2 & 9.4 & 25.5 \\
            no v-len normalization & 73.2 & 69.8 & 36.6 & 3.6 & 14.4 & 7.4 & 2.0 & 2.2 & 12.6 & 5.2 & 0.2 & 2.6 & 5.3 & 13.4 \\
            no merge gate & 95.2 & 91.6 & 88.0 & 87.8 & 33.8 & 9.8 & 2.6 & 2.4 & 28.4 & 4.0 & 3.0 & 0.6 & 12.1 & 20.3 \\
            \bottomrule
        \end{tabular}
    \end{adjustbox}
    \caption{NIAH, RULER-4096 and average of LongBench ("LB") few-shot evaluations for KVM ablations.}
    \label{tab:ablations-evals-long}
\end{table}

The ablations show that our architectural choices primarily affect long context behavior. Removing value-length normalization leads to the largest degradation, while removing sink protection and the merge gate also substantially weaken long-context retrieval.

\section{Conclusions} \label{conclusion}

We introduced Key-Value Means (KVM), an attention mechanism that consists of block sliding-window attention and an expandable compressive state in a single softmax attention layer. It provides a flexible choice of state size, unlike fixed-size RNNs and full-attention transformers. With fixed state, it provides an $O(N)$ chunked recurrent architecture, and with growable state it recovers substantially stronger long-context behavior with sublinear asymptotic state growth. KVM exhibits competitive short-context performance and has strong long-range retrieval, tunable using different state-size schedules. KVM shows that, instead of choosing between fixed-state RNNs and full attention, it is possible to interpolate between them smoothly in a simple and effective manner.

\paragraph{Future Work} In our experiments, we trained KVM on static schedules for state size/chunk size; it may be of interest to change different aspects of such scheduling - changing scheduling between train/test time, scheduling adaptation via finetuning, data-dependent scheduling, and so on. We have not yet tried standard methods of improving transformer parameter and KV cache efficiency such as GQA \citep{ainslie2023gqa}, MLA \citep{deepseekai2024deepseekv2strongeconomicalefficient}, etc. but we believe they should apply easily and directly to KVM.

We believe it may be possible to efficiently distill transformers to use KVM attention on one or more layers, thereby reducing their memory footprint and other costs. Although we have not yet attempted this, the query, key and value projection seem very likely to align closely with a teacher model because KVM uses traditional attention and even attends to a BSWA window with no special changes beyond a simple temperature adjustment. We leave exploration of this promising direction to future work.

\section*{AI Usage Disclosure}

We used LLMs to help with code and math tasks, to generate diagrams as TikZ code and to suggest phrasing and stylistic improvements for this paper. We also discussed mathematical and code topics with LLMs during our research process, and improved our coverage of relevant literature using LLM-based search tools.

\newpage

\bibliography{main}
\bibliographystyle{colm2026_conference}

\newpage
\appendix

\section{Related Work} \label{sec:related-work}

The use of state, also known as fast weights \citep{schmidhuber1992learning,schlag2021linear} to train an inner model at test time can be a very powerful concept, allowing models to learn and grow not just through pretraining but based on user input. RNN state is a form of fast weights, and even attention itself can be viewed as a set of expanding fast weights. It has recently become common to take the idea of training fast weights literally, using classic optimizers like SGD, Adam or even newer ones like Muon at runtime. Speed is a challenge with such techniques. KVM is positioned within this broader landscape but avoids runtime optimizers and their associated hyperparameters, relying instead upon a simple state update rule.

\paragraph{Fixed-Size State Architectures} There have been many architectures that feature a fixed-size state, which come in both linear and nonlinear varieties. These models provide attractive fixed memory cost and fixed amortized computation per token during inference, but face challenges with retrieval over long contexts as their total memory is necessarily limited.

Block-Recurrent Transformers (BRT) \citep{NEURIPS2022_d6e0bbb9} apply a block-wise recurrence to periodically update a fixed-size state. A Sliding Window Attention (SWA) pass over its input token stream is concatenated with a cross-attention pass over the state, and projected. Its state recurrence is self-attention over the state with cross attention over the incoming block of input tokens, which is then gated. BRT requires an extra set of projection matrices dedicated to its state, using more parameters than an equivalent transformer. TransformerFAM \citep{hwang2024transformerfamfeedbackattentionworking} extends this by using Block Sliding Window Attention (BSWA) and eliminating the extra projections, instead employing the existing FFN to reformat its state output. Crucially, it compresses the overflow from BSWA into its state after every chunk.

Native Hybrid Attention (NHA) \citep{nativehybridattention2026} maintains a constant number of long term key-value slots and concatenates them with tokens from a sliding window, to attend to them using a single softmax, with RoPE applied only to the local sliding window tokens. As tokens leave the local window, they are then merged into the long-term state. This shares KVM's use of a single softmax across its short-term and long-term memory, but NHA uses a constant number of smoothly updated recurrent slots (using a linear RNN) rather than a winner-take-all assignment into an expandable state.

Linear attention \citep{katharopoulos2020lineartransformrers} variants, state space models, and LRNNs in general typically employ a fixed-size state, with a simple update rule that can be efficiently parallelized across the time dimension \citep{yang2024parallelizing}, at least over short chunks. Modern variants like RWKV-7 \citep{peng2025rwkv7gooseexpressivedynamic}, Gated DeltaNet (GDN) \citep{yang2025gated}, and Kimi Delta Attention (KDA) \citep{kimiteam2025kimilinearexpressiveefficient} use a matrix-valued state with an Identity Plus Low Rank (IPLR) or Diagonal Plus Low Rank (DPLR) update rule, which directly implements a form of gradient descent. This typically requires a custom kernel for high-speed training and inference.

Test-Time Training (TTT) \citep{sun2025learning} layers treat the state as the weights of a shallow neural network and update it via mini-batched gradient descent during inference. This perspective on training fast weights at test time has led to a series of architectures that expand upon and generalize the core idea.

Titans \citep{behrouz2025titans} separates fixed-size state into 1) Core, 2) Long-Term Memory (LTM), and 3) Persistent Memory, and identifies three generalized implementation strategies for models with such LTM components: i) Memory As Context (MAC), ii) Memory As Layer (MAL), or iii) Memory As Gated branch (MAG). Their core is always attention, but it can attend to token sub-segments generated in various ways. Their LTM takes models like GDN and RWKV-7 and generalizes them from single-layer matrix state to all possible nonlinear simple MLPs with one or more layers. In order to enable chunked parallelization despite having a nonlinear recurrence, they treat the state update as mini-batched gradient descent. In this way, it is a generalization of TTT. Their Persistent Memory consists of a learned prefix that is prepended to their current context segment. Unfortunately, their models are still slow to train and slow at inference time.

Much like the Titans LTM, Large Chunk Test-Time Training (LaCT) \citep{zhang2026testtime} employs nonlinear fast weights set up as a two-layer SwiGLU-MLP, and uses classic backpropagation with the Muon optimizer and momentum as the update rule. To reduce the computational burden of this complex update rule, they batch larger updates every 2048 tokens or more. This permits fast inference and training per token, but has the downside that training requires fairly long contexts. They integrate this with SWA via a form of MAG.

\paragraph{Expandable State Size Architectures} In a reflection of the difficulties with expanding weights during pretraining, a smaller body of work considers architectures whose fast-weight state grows over time. This may seem somewhat surprising, as attention itself expands its fast weights at test time through a growing key-value cache. A key challenge has been in growing state more slowly than full attention while still allowing capacity to increase over time, while maintaining high-quality results. 

Compressive Transformer \citep{Rae2020Compressive} takes blocks that overflow from a BSWA window and compresses them by a fixed ratio using one of several methods, e.g. convolution. These compressed blocks are then added to a FIFO queue. Attention is performed uniformly across both compressed blocks in the FIFO queue and uncompressed tokens in the BSWA window.

Concurrent work End-to-End Context Compression at Scale \citep{latentcontextlms2026} trains an encoder that maps fixed-size blocks of raw tokens into continuous latent tokens, that are then processed by the decoder. The number of latent tokens hence continues to grow with the context length, but at a reduced rate that is determined by the compression ratio of the encoder. Unlike KVM, its context is compressed by a separate encoder rather than being maintained as a compressed recurrent state by the model itself.

TokenFormer \citep{wang2025tokenformer} considers a two-layer MLP that mimics the Key-Value Cache from standard attention, but with a revised version of softmax that admits the ability to dynamically expand this state size without changing its outputs. Their focus is using this to expand weights (and hence, scale model size) during pretraining. As such, they do not directly experiment with applying this method to attention itself, but consider it for future work.

Online Vector Quantization (OVQ) \citep{alonso2026onlinevectorquantizedattention} maintains a capped-size dictionary of quantized key-value centroids that are updated as a running average of the best-matching incoming tokens. It is a layerwise hybrid with sliding window attention, relying on the sliding window layers for positional encoding of short-context information.

Concurrent with our work, OVQ shares a winner-take-all assignment strategy with KVM. The main differences are that KVM (1) integrates compressed state and BSWA attention in a single softmax pass rather than separate layers, (2) does not require per-centroid count tracking due to renormalization and includes additional dynamic weighting, (3) addresses RoPE compatibility explicitly via partial-dimension zeroing, (4) supports uncapped state expansion, (5) is sink-aware through preserving sinks as well as value magnitudes, and (6) separates the state and BSWA regions via learned softmax temperatures.

Many recent KV compression techniques similarly merge multiple tokens into a smaller set of key-value clusters/centroids. Concurrent with our work, SemantiCache \citep{semanticache2026} clusters similar keys within semantic chunks of tokens and replaces each cluster with key-value centroids while applying a correction to their attention logits according to the number of tokens in the cluster. CentroidKV \citep{centroidkv2026} uses chunked similarity-based matching to identify groups of key-value pairs to merge into clusters/centroids. ZSMerge \citep{liu2025zsmergezeroshotkvcache} divides a cache budget into recent, important, and residual regions, with evicted entries merged into the most similar residual centroid and given a count-dependent logit correction. These techniques are typically training-free unlike KVM, and explicitly track/compensate for the number of tokens in the cluster represented by each centroid, in order to more closely approximate attention, unlike KVM's JIT renormalization, which does not try to approximate attention.

\section{Pseudocode}
\label{sec:pseudocode}

\begin{verbatim}
# Pseudocode for chunk state update recurrence with attention output
def inner_loop_attstate(self, x, q, k, v, s_k, s_v, s_vlen, bswa_begin, bswa_end, sink_len):
    # identify overflow chunk of tokens to merge into (or append to) the state
    o_k = k[:,:,bswa_begin-chunk_len:bswa_begin]
    o_v = v[:,:,bswa_begin-chunk_len:bswa_begin]

    # note: some tokens out of these will be appended, split and append
    # to be done as explained in the main text
    
    # remove rope and apply data-dependent weighting to the tokens to be merged
    g = 1 + elu(x @ self.W_merge_gate)[:,:,bswa_begin-chunk_len:bswa_begin]
    o_k = self.layernorm_s_k(remove_rope(o_k)) * g
    o_v = o_v * g
    
    # obtain normalized state keys
    s_k_norm = self.layernorm_s_k(s_k)
    
    # find the most similar key in state for each overflow key to merge
    logits = o_k @ s_k_norm.mT
    # avoid protected sinks
    logits[...,0:sink_len] = float('-inf')
    best_s_idx = logits.max(dim=-1, keepdim=True).indices
    scores = scatter(zeros_like(logits), -1, best_s_idx, torch.ones_like(logits))
    
    # update state by adding the most similar keys and their values
    s_k = s_k + (scores.mT @ o_k)
    s_v = s_v + (scores.mT @ o_v)
    
    # calculate attention across the newly updated state and BSWA window
    a_q = q[:, :, bswa_end-chunk_len:bswa_end]
    s_k_attn = self.layernorm_s_k(s_k) * self.state_temperature
    bswa_k = k[:, :, bswa_begin:bswa_end] * self.bswa_temperature
    s_v_attn = (normalize(s_v.float(), dim=-1) * s_vlen).to(s_v.dtype)
    bswa_v = v[:, :, bswa_begin:bswa_end]
    a_k = cat([s_k_attn, bswa_k], dim=-2) 
    a_v = cat([s_v_attn, bswa_v], dim=-2)
    out = sdpa(a_q, a_k, a_v, attn_mask=causal_mask_after_state)
    
    return s_k, s_v, out
\end{verbatim}

Please note that the recurrence alternates with attention in the pseudocode above, but it does not have to be implemented this way. The state recurrence can be calculated with its results stored, and then a single call to attention masked to operate across both the block sliding window and related state regions suffices for training or prefill, e.g. using FlexAttention \citep{dong2024flexattentionprogrammingmodel}. Also, the pseudocode processes the state first (using lagging information from the chunk that falls off the previous loop iteration) and then does attention, for the sake of simplicity; this is semantically equivalent to the main text's description.

\section{GPTAlpha-2 Transformer Architecture and Backbone}
\label{sec:gptalpha}

For the experiments in this paper we use a modified version of the GPTAlpha transformer architecture found in \cite{goldfinch}, incorporating several design choices from RWKV-7. We call this GPTAlpha-2. This includes LayerNorm with bias on queries and keys with a simplified non-data-dependent token shift, value residuals \citep{zhou2025valueresiduallearning}, and RoPE. Unless otherwise noted, we apply RoPE across only half of the dimension of each head.

For the channel mixing MLP, we use the RWKV-7 channel mixer.

\paragraph{GPTAlpha-2 Attention weight preparation} (single head shown for notational convenience)
\begin{align}
\tilde{\mathbf{q}}_t &= \mathbf{x}_t W_Q, &\text{simple query}\\
\tilde{\mathbf{k}}_t &= \mathbf{x}_t W_K, &\text{simple key}\\
\tilde{\mathbf{v}}_t &= \mathbf{x}_t W_V, &\text{simple value}\\
\tilde{\mathbf{v}}_t &
\leftarrow
(1-\lambda)\tilde{\mathbf{v}}_t + \lambda \tilde{\mathbf{v}}_{t}^{\mathrm{first}}, \quad \lambda\in\mathbb{R}^{d_h} &\text{value residual}\\
\mathbf{a}_t&=
\tilde{\mathbf{a}}_t+\boldsymbol{\alpha}_a\odot(\tilde{\mathbf{a}}_{t-1}-\tilde{\mathbf{a}}_t),
\qquad
\mathbf{a}_0=\tilde{\mathbf{a}}_0, \qquad\mathbf{a}\in\{\mathbf{q},\mathbf{k},\mathbf{v}\} &\text{token shift}\\
\mathbf{q}_t& \leftarrow\operatorname{RoPE}_r(\operatorname{LN}_q(\mathbf{q}_t)),
&\text{RoPE query}\\
\mathbf{k}_t& \leftarrow\operatorname{RoPE}_r(\operatorname{LN}_k(\mathbf{k}_t)).
&\text{RoPE key}
\end{align}

where $\tilde{\mathbf{v}}_{t}^{\mathrm{first}}$ is the $\tilde{\mathbf{v}}_t$ calculated for the first layer.

\paragraph{GPTAlpha-2 Channel Mixer}
\begin{align}
\mathbf{h}_{t}&=
(\mathbf{x}_t+\boldsymbol{\alpha}\odot(\mathbf{x}_{t-1}-\mathbf{x}_t)) W_U, \quad \boldsymbol{\alpha}\in\mathbb{R}^d
 &\text{intermediate hidden state}\\
\mathbf{o}_{t} &= \operatorname{ReLU}(\mathbf{h}_{t})^2 W_D. &\text{output}
\end{align}

\section{Training details and Hyperparameters}
\label{sec:hyperparameters}

We use CompleteP \citep{dey2025dont} with $\alpha = 1$ for parameter-wise depth and width scaling. For longer runs, we scale the learning rate and weight decay by $\frac{1}{\sqrt{N_{\text{steps}}}}$ (where $N_{\text{steps}}$ is the total number of training steps), while keeping batch size constant at 524,288 tokens.

We use the AdamC optimizer \citep{defazio2025gradientsrapidlyincreasenear} for weight decay scheduling. This is expected to keep parameter norms stationary over long training and improves performance for us as compared to AdamW. We use $\beta_1 = 0.9$, $\beta_2 = 0.95$, $\epsilon = 10^{-8}$, base learning rate tuned to $2 \times 10^{-3}$ and a weight decay tuned to $0.2$.

Learning rates and weight decay were tuned for a 120M model with 3B tokens, and then transferred to larger scales. We do not apply weight decay to scalar/vector parameters.

We use a warmup of 200 steps, then a linear decay to 0 \citep{bergsma2025straightzerolinearlydecaying} for the learning rate for the rest of the steps.

We set the RoPE base to 10,000, even when applying across only 64 of 128 channels.

\section{Short-context evals}
\label{sec:evals-short}

Table~\ref{tab:evals-short} contains short-context evaluation results for our models as reported by LM Evaluation Harness for each evaluation.

Table~\ref{tab:ablations-evals-short} contains short-context evaluation results for KVM ablations, and Table~\ref{tab:evals-hrope-short} contains short-context evaluation results for GPTA-2 partial RoPE ablations.

We abbreviate LAMBADA \citep{paperno2016lambada} as lmbda, ARC-Challenge \citep{clark2018thinkarc} normalized as arc\_c, ARC-Easy as arc\_e, HellaSwag \citep{zellers2019hellaswagmachinereallyfinish} normalized as hella, PIQA \citep{bisk2020piqa} as piqa, and WinoGrande \citep{sakaguchi2021winogrande} as winog.
\begin{table}[htb]
    \centering
    \begin{adjustbox}{max width=\linewidth}
        \begin{tabular}{lrrrrrrrr}
            \toprule
            Architecture & lmbda ppl$\downarrow$ & lmbda$\uparrow$ & arc\_c$\uparrow$ & arc\_e$\uparrow$ & hella$\uparrow$ & piqa$\uparrow$ & winog$\uparrow$ & avg.$\uparrow$ \\
            \midrule
            120M BSWA & 47.7 & 31.5 & 24.5 & 50.1 & 32.6 & 63.5 & 50.4 & 42.1\\
            120M RWKV-7 & 42.6 & 31.7 & 24.7 & 49.8 & 33.6 & 64.3 & 52.2 & 42.7 \\
            120M GPTA-2 & 51.5 & 31.0 & 24.7 & 49.2 & 32.8 & 64.4 & 50.0 & 42.0 \\
            120M KVM 256 & 53.4 & 30.3 & 23.7 & 49.8 & 33.1 & 64.1 & 51.0 & 42.0 \\
            120M KVM sqrt & 51.0 & 31.0 & 25.1 & 51.1 & 33.1 & 64.1 & 51.6 & 42.7 \\
            \midrule
            120M OVQ/SWA & 57.3 & 29.9 & 26.1 & 49.5 & 32.4 & 63.4 & 50.1 & 41.9 \\
            120M GPTA-2 HalfRoPE/SWA & 51.6 & 31.0 & 23.5 & 50.2 & 33.5 & 64.7 & 50.0 & 42.1 \\
            120M GPTA-2 NoPE/SWA & 56.5 & 29.7 & 24.0 & 49.5 & 32.8 & 64.1 & 51.9 & 42.0 \\
            120M KVM/SWA & 51.8 & 30.6 & 24.7 & 49.6 & 33.1 & 63.5 & 52.1 & 42.3 \\
            \midrule
            350M BSWA & 22.2 & 38.8 & 27.0 & 56.4 & 41.9 & 68.9 & 51.9 & 47.5 \\
            350M RWKV-7 & 19.9 & 40.2 & 26.9 & 57.1 & 42.9 & 69.3 & 53.0 & 48.2 \\
            350M GPTA-2 & 22.9 & 38.2 & 27.1 & 56.2 & 41.7 & 68.2 & 51.5 & 47.2 \\
            350M KVM 256 & 22.0 & 38.2 & 27.0 & 56.9 & 41.9 & 68.6 & 51.3 & 47.3 \\
            350M KVM sqrt & 21.7 & 38.8 & 27.0 & 57.0 & 42.0 & 69.1 & 52.2 & 47.7 \\
            \midrule
            350M OVQ/SWA & 23.2 & 38.8 & 25.4 & 56.9 & 41.5 & 67.7 & 51.4 & 46.9 \\
            350M GPTA-2 HalfRoPE/SWA & 21.6 & 38.9 & 27.2 & 57.0 & 42.3 & 69.0 & 50.7 & 47.5 \\
            350M GPTA-2 NoPE/SWA & 22.9 & 38.6 & 27.6 & 57.1 & 41.7 & 68.0 & 51.1 & 47.4 \\
            350M KVM/SWA & 22.0 & 38.8 & 26.9 & 57.7 & 42.7 & 68.6 & 49.4 & 47.3 \\
            \bottomrule
        \end{tabular}
    \end{adjustbox}
    \caption{Standard short-context language modeling evaluations}
    \label{tab:evals-short}
\end{table}

\begin{table}[htb]
    \centering
    \begin{adjustbox}{max width=\linewidth}
        \begin{tabular}{lrrrrrrrr}
            \toprule
            Architecture & lmbda ppl$\downarrow$ & lmbda$\uparrow$ & arc\_c$\uparrow$ & arc\_e$\uparrow$ & hella$\uparrow$ & piqa$\uparrow$ & winog$\uparrow$ & avg.$\uparrow$ \\
            \midrule
            baseline & 53.4 & 30.3 & 23.7 & 49.8 & 33.1 & 64.1 & 51.0 & 42.0 \\
            no sink & 56.3 & 28.7 & 24.7 & 49.0 & 32.8 & 64.6 & 51.7 & 41.9 \\
            no head temps & 53.4 & 30.1 & 24.1 & 51.6 & 33.0 & 63.1 & 51.1 & 42.2 \\
            no v-len normalization & 53.3 & 30.1 & 24.5 & 50.5 & 32.8 & 65.2 & 52.6 & 42.6 \\
            no merge gate & 51.0 & 30.6 & 25.1 & 50.1 & 32.9 & 65.1 & 50.9 & 42.5 \\
            \bottomrule
        \end{tabular}
    \end{adjustbox}
    \caption{Standard short-context language modeling evaluations for KVM ablations.}
    \label{tab:ablations-evals-short}
\end{table}

\section{Extrapolation and partial RoPE ablations}
\label{sec:extrapolation}

We observed that using NoPE and HalfRoPE (i.e., NoPE on half the dimensions and RoPE on the other half) for hybrid GPTA-2 models had materially different results when it came to length extrapolation. On position-wise loss for the TextbookChapters dataset, we see that NoPE has increasing loss values, while HalfRoPE has stable loss values. For long context evaluations (NIAH/LongBench/RULER), HalfRoPE generally outperforms NoPE within the context length, but is typically worse as further extrapolation occurs. We also observe the effect of training length for the NoPE model - more training worsens the out-of-the-box length extrapolation capabilities of the NoPE model as measured by per-position loss, while improving its NIAH/LongBench/RULER scores.

One possible explanation for this is the following: vanilla NoPE tends to learn absolute positional embeddings \citep{haviv-etal-2022-transformer,kazemnejad2023impactpositionalencodinglength}, and as the amount of training (at a fixed training context length) increases, how strongly the model relies on these absolute position representations learned by NoPE also increases. This may make vanilla NoPE variants less suitable than HalfRoPE variants for extrapolating beyond the training context length in some aspects. OVQ also relies on NoPE in its compressed state, which may contribute to its weaker position-wise length extrapolation relative to KVM.

We conjecture that NoPE, while being bad at extrapolation in terms of loss, focuses more on global aspects of long-context modeling and succeeds at pinpointing specific tokens. However, it appears to be weaker at NIAH-S3, potentially since it is expected to attend to multiple tokens adjacent to each other, and HalfRoPE has stronger short-context/relative-position handling capabilities.

\begin{figure}
    \centering
    \includegraphics[width=1\linewidth]{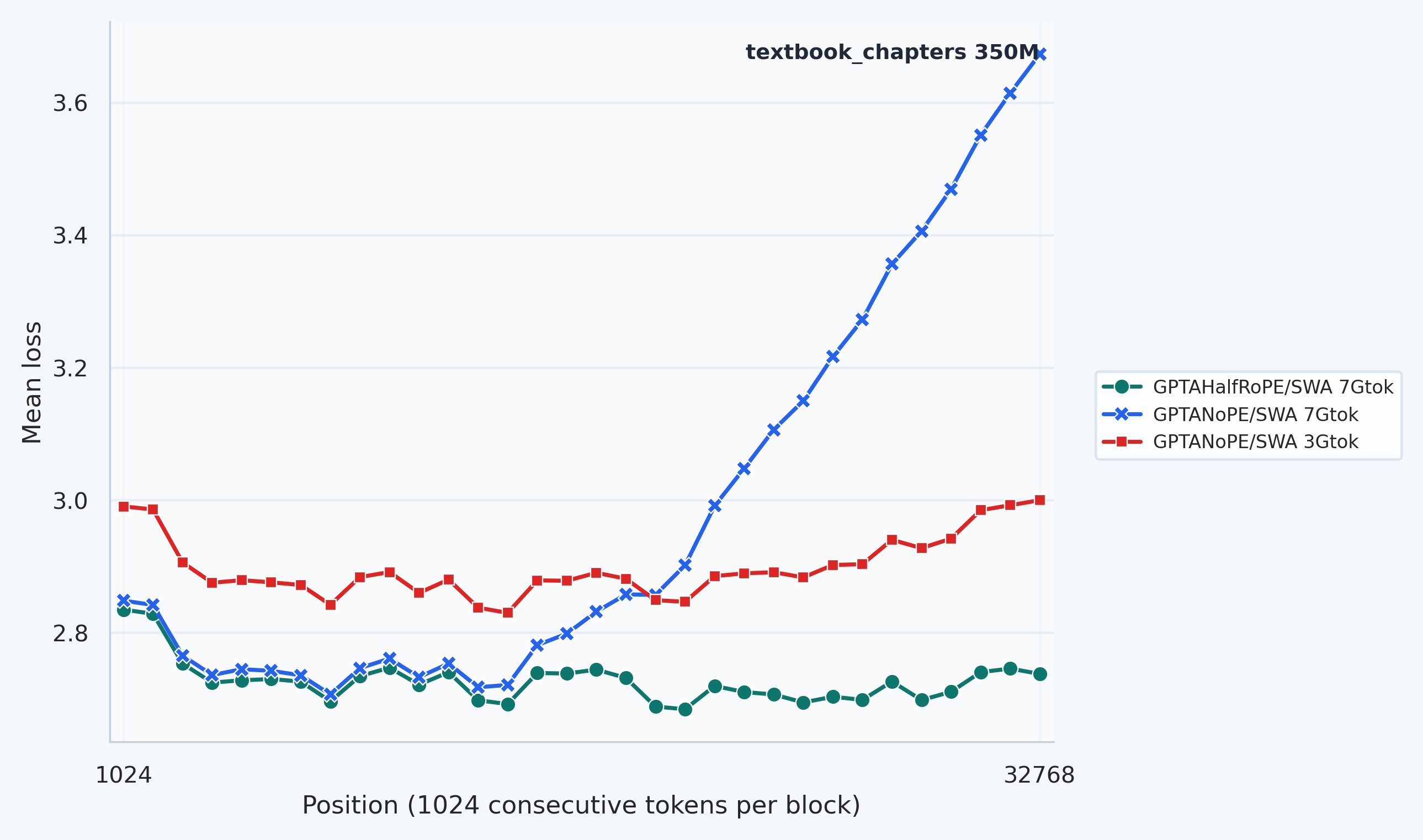}
    \caption{TextbookChapters GPTAlpha-2/SWA mean loss per 1024 token block}
    \label{fig:textbook-chapters-hrope}
\end{figure}

Figure~\ref{fig:textbook-chapters-hrope}, Table~\ref{tab:evals-hrope-long}, and Table~\ref{tab:evals-hrope-short} illustrate these results.

\begin{table}[htb]
    \centering
    \begin{adjustbox}{max width=\linewidth}
        \begin{tabular}{l*{14}{r}}
            \toprule
            & \multicolumn{4}{c}{NIAH-S1$\uparrow$}
            & \multicolumn{4}{c}{NIAH-S2$\uparrow$}
            & \multicolumn{4}{c}{NIAH-S3$\uparrow$} & LB$\uparrow$ & RULER$\uparrow$ \\
            \cmidrule(lr){2-5} \cmidrule(lr){6-9} \cmidrule(lr){10-13}
            Architecture & 4K & 8K & 16K & 32K & 4K & 8K & 16K & 32K & 4K & 8K & 16K & 32K & avg. & avg. \\
            \midrule
            HalfRoPE 7.8Gtok & 99.2 & 99.6 & 92.6 & 41.2 & 99.4 & 99.4 & 52.0 & 23.6 & 88.2 & 60.4 & 37.2 & 9.6 & 13.6 & 41.6 \\
            NoPE 7.8Gtok & 99.4 & 99.6 & 99.6 & 95.0 & 99.4 & 97.6 & 99.0 & 0.0 & 82.2 & 47.8 & 29.2 & 0.0 & 21.0 & 45.1 \\
            NoPE 3Gtok & 100.0 & 99.8 & 100.0 & 99.4 & 98.4 & 96.2 & 53.6 & 2.4 & 73.4 & 37.6 & 10.6 & 1.8 & 6.0 & 40.3 \\
            \bottomrule
        \end{tabular}
    \end{adjustbox}
    \caption{NIAH, RULER-4096 and average of LongBench ("LB") few-shot evaluations for 350M GPTA-2/SWA hybrid RoPE ablations}
    \label{tab:evals-hrope-long}
\end{table}

\begin{table}[htb]
    \centering
    \begin{adjustbox}{max width=\linewidth}
        \begin{tabular}{lrrrrrrr}
            \toprule
            Architecture & lmbda ppl$\downarrow$ & lmbda$\uparrow$ & arc\_c$\uparrow$ & arc\_e$\uparrow$ & hella$\uparrow$ & piqa$\uparrow$ & winog$\uparrow$ \\
            \midrule
            HalfRoPE 7.8Gtok & 21.6 & 38.9 & 27.2 & 57.0 & 42.3 & 69.0 & 50.7 \\
            NoPE 7.8Gtok & 22.9 & 38.6 & 27.6 & 57.1 & 41.7 & 68.0 & 51.1 \\
            NoPE 3Gtok & 32.9 & 34.0 & 25.9 & 54.0 & 36.9 & 66.5 & 49.6 \\
            \bottomrule
        \end{tabular}
    \end{adjustbox}
    \caption{Standard short-context language modeling evaluations for 350M GPTA-2/SWA hybrid RoPE ablation}
    \label{tab:evals-hrope-short}
\end{table}

\section{Design choices}\label{sec:design-rationale}

\paragraph{Motivation}

Our goal is a high-performance new long-context centric architecture that has constant or sublinear memory growth and subquadratic computational complexity with respect to sequence length. To this end, we seek a growable compressive state architecture that is efficient and high-quality, and minimizes the need for hyperparameters that control its test-time training.

\paragraph{Overall BSWA framework}

Traditional softmax attention is the standard for transformers over long contexts, making it a leading candidate for inclusion in this architecture. Therefore we would like our state to contain entries for both keys and values so that we can perform attention across these. Traditional attention appends to the state at each token, but our goal is to grow less quickly than that, forcing us to update the state in-place at least some of the time. Batching is a straightforward way to increase efficiency given the parallel nature of modern GPU architectures, so we process our state updates in batches of many tokens. But batching implies that some tokens will remain un-integrated into the state until a batch is full. Fortunately, BSWA provides a natural mechanism for attending to these as yet un-integrated tokens, since the compression step can easily occur at the time of the change in window size: when a block overflows the window and is removed from view, we compress that block's information into the state. We attend to the concatenation of the BSWA window and the separate compressed state.

\paragraph{State Compression}

We now have a candidate for the overall framework, but we still require compatible high-quality methods of compression and state expansion. We tackle compression first, holding state size fixed for the moment. Notice that calculating an attention matrix of attention logits between the overflow keys and the state keys provides a natural way to determine how much of each overflow key to compress into each state key, based on their mutual similarity. Traditional attention would apply softmax to these logits to obtain the final metric for an overflow key-state key pair, but there exist many other possibilities.

We consider many alternatives for this metric, including various $\phi$ functions of the logits as in classical linear attention, deferred normalization as seen in modern LRNNs, all possible $L_n$ normalizations of these logits up through $L_{\infty}$ as in many modern LRNNs, and variations on softmax attention employing different temperatures and normalizations and exponentiations. (The $L_1$ normalization of the exponentiated logits gives the traditional attention scores.) Experimentally, performance improved as we decreased temperature or exponentiated further. In the limit this is equivalent to an attention matrix containing 1.0 at the maximum logit from each row and 0 for all others. OVQ made this choice, and inspired us to increase the range of our normalization attempts, which improved our results significantly. One possible explanation is that maximizing the distance between state keys would preserve separability, allowing more information to be stored successfully, motivating such a maximally sparse update matrix.

We have now determined generally how much of each overflow key-value pair should be merged into each state key-value pair. But the exact method of the merger is still undecided. Potential choices include whether to keep a running average or an exponential moving average, whether to weight the incoming overflow token, whether to first decay the pre-existing state token in either a simple or delta-rule like fashion, and whether to renormalize the merge result. Renormalization is convenient as it eliminates the need to separately track totals for each token for averaging purposes, but there is also a strong mathematical reason to prefer renormalization: when averaging multiple vectors together, orthogonal input vectors cause a reduction in norm of the average of the vectors, and opposing components of input vectors cause destructive interference, further reducing the norm of the average of those vectors. So in order to avoid KV vectors that shrink over time, we must renormalize just-in-time (JIT norm) prior to attention.

Experiments showed that keeping a running average outperformed EMA, that weighting the incoming overflow token was important, and that our hypothesis about JIT norm was important. Because query/key normalization is often used to improve attention and has theoretical motivations from test-time regression \citep{wang2025testtimeregressionunifyingframework}, it makes sense that we should apply that same norm as a JIT norm to our state keys. This allows us to keep the state keys as a simple sum of weighted incoming overflow keys. The remaining design choice is how to treat state values. We find that the norm of our values is important, and that sink tokens can have very different norms than other tokens \citep{guo2024attentionscoreneedtoken}. To avoid overspecializing our architecture, we simply take the initial norm of each starting state value, store that, and use it as the JIT norm for that state value for the lifetime of that state row. This works well in practice, while allowing each state value to be JIT normalized to its own unique radius.

\paragraph{State Initialization and Expansion}

A natural expansion rule is to append the most surprising overflow tokens, i.e. the least redundant ones under the current state similarity metric. If we start out our sequence imagining that there is no state at all then we are presented with a convenient opportunity to define this expansion inductively. At the first state-creation step, the overflowing tokens are by definition the most surprising, and we can simply initialize the state with these tokens. This implies a similar strategy for future overflow tokens; we can simply append the most surprising ones to the state, and then merge the remaining overflow tokens into this newly expanded state. We may choose a similarity threshold for this expansion condition as a hyperparameter, as a learned value according to some loss metric, or simply choose a fixed schedule at which to expand the state size. For simplicity, we choose a fixed schedule and leave a learned value cutoff to future work.

\paragraph{Positional Encoding}

We still need a way to deal with positional encoding of the state. There is a recent trend towards using NoPE on long context layers, and RoPE on short context layers \citep{yang2025rope}. Since our state never encodes the short context in BSWA, and because the key positions may come to encompass keys from widely varying positions in the set of overflow windows, it is natural to avoid RoPE in the state. But the question remains of how to do so without sacrificing downstream performance or requiring extra parameters. Several options are available, including artificially placing all state keys at a specific fixed RoPE sequence position, separating the attention over the state from that over the BSWA window and re-merging these using logsumexp outputs so that we can use unrotated queries and state keys for the state but RoPE on the BSWA window, or using partial RoPE and zeroing the RoPE portion of the state keys, or zeroing the rotational subspace of the keys before merging them into the state. For simplicity we never tried the attention re-merging mechanism, but it seems promising and we leave it and other options to future work. The last mechanism works well for us in practice, but we believe there is more downstream performance not captured by this design choice since it removes some expressivity from some of our state key dimensions.

\end{document}